\documentclass[preprint,12pt]{elsarticle}

\usepackage{graphicx}
\usepackage{amsmath}
\usepackage{graphicx}
\usepackage{lipsum}
\usepackage{rotating}
\usepackage{hyperref}
\usepackage{xcolor}
\usepackage{lineno}
\usepackage{pifont}
\newcommand{\cmark}{\ding{51}}
\newcommand{\xmark}{\ding{55}}
\usepackage{array}
\usepackage{multirow}
\newcolumntype{?}{!{\vrule width 2pt}}
\date{}

\begin{document}
	
\title{
Leveraging Medical Foundation Model Features in Graph Neural Network-Based Retrieval of Breast Histopathology Images
}
	
\author[1]{Nematollah Saeidi}
\ead{saeidi.n@eng.ui.ac.ir}
\author[1]{Hossein Karshenas}
\ead{h.karshenas@eng.ui.ac.ir}
\author[1,2]{Bijan Shoushtarian\corref{cor1}}
\ead{shoushtarian@eng.ui.ac.ir, bshoushtarian@mpu.edu.iq}
\author[3,4]{Sepideh Hatamikia}
\ead{sepideh.hatamikia@dp-uni.ac.at}
\author[5]{Ramona Woitek}
\ead{ramona.woitek@dp-uni.ac.at}
\author[5]{Amirreza Mahbod}
\ead{amirreza.mahbod@dp-uni.ac.at}

\cortext[cor1]{Corresponding author}
\address[1]{Artificial Intelligence Department, Faculty of Computer Engineering, University of Isfahan, Isfahan, Iran}
\address[2]{Department of Computer Engineering Techniques, Mazaya University College, Nasiriyah, Iraq}
\address[3]{Department of Medicine, Danube Private University, Krems an der Donau, Austria}
\address[4]{Austrian Center for Medical Innovation and Technology, Wiener Neustadt, Austria}
\address[5]{Research Center for Medical Image Analysis and Artificial Intelligence, Department of Medicine, Danube Private University, Krems an der Donau, Austria}
	
	
\begin{abstract}
Breast cancer is the most common cancer type in women worldwide. Early detection and appropriate treatment can significantly reduce its impact. While histopathology examinations play a vital role in rapid and accurate diagnosis, they often require 
experienced medical experts for proper recognition and cancer grading. Automated image retrieval systems have the potential to assist pathologists in identifying cancerous tissues, thereby accelerating the diagnostic process. Nevertheless, proposing an accurate image retrieval model is challenging due to considerable variability among the tissue and cell patterns in histological images. 

In this work, we leverage the features from foundation models in a novel attention-based adversarially regularized variational graph autoencoder model for breast histological image retrieval. Our results confirm the superior performance of models trained with foundation model features compared to those using pre-trained convolutional neural networks (up to 7.7\% and 15.5\% for mAP and mMV, respectively), with the pre-trained
general-purpose self-supervised model for computational pathology (UNI) delivering the best overall performance. By evaluating two publicly available histology image datasets of breast cancer, our top-performing model, trained with UNI features, achieved average mAP/mMV scores of 96.7\%/91.5\% and 97.6\%/94.2\%  for the BreakHis and BACH datasets, respectively.


Our proposed retrieval model has the potential to be used in clinical settings to enhance diagnostic performance and ultimately benefit patients. 	
\end{abstract}
	
\maketitle
	
\section{Introduction}
Breast cancer is a global health concern for women and has led to considerable mortality worldwide~\cite{Rashmi2021}. It is the most commonly diagnosed cancer and represents around 25\% of all cancers diagnoses in women~\cite{ARNOLD202215}.  According to 2018 statistics, breast cancer accounted for 15\% of all cancer-related deaths and has increased by 24\% in recent years~\cite{Minarno2021}. While breast cancer causes many deaths worldwide, early detection and diagnosis, followed by appropriate treatment, can help to reduce its impact~\cite{Burstein2019}. The need for early detection of breast cancer has led to new solutions, such as computer-aided diagnosis systems based on histological images~\cite{Tabatabaei2022}. This technology is a valuable tool to improve breast cancer diagnosis and potentially can save lives~\cite{Fuster2022}. However, proper diagnosis based on histological images often requires a substantial workforce and experienced medical experts. Pathologists can benefit from referring to archived databases for complex cases to find similar examples with verified findings to minimize diagnostic errors and overcome this challenge. 

Content-based image retrieval (CBIR) is an approach to identifying visually similar images in a database and can be used for various applications~\cite{10.1007/978-981-16-8892-8_33, 8594636}. 
Content-based medical image retrieval (CBMIR), a subset of CBIR, is specifically adapted to medical images and has been used in the context of radiology and histology in former studies~\cite{Z2018, Silva-Rodriguez2020}. 
The CBMIR of histological images not only aids histopathology in grading new tissue samples but also assists in analyzing patterns within similar tissues from previous patients~\cite{Silva-Rodriguez2020, Tabatabaei2022}. 

Histopathology image retrieval is challenging due to the substantial tissue and cell pattern variability~\cite{Hegde2019}. Various semi- and fully-automatic approaches have been proposed for histopathology image retrieval~\cite{Z2018}. Among these, graph neural networks (GNN) and  graph convolutional neural networks (GCNN) have gained considerable attention in recent years~\cite{Yamashita2018}. 
While supervised learning is usually used to train GCNNs, some GCNN variants, such as variational graph autoencoder (VGAE)~\cite{Xiong2021}, can be used for unsupervised learning tasks. 
VGAE performance can be further improved by incorporating other techniques, such as adversarial training or adding an attention mechanism to the model~\cite{Xiong2021}. In an adversarial regularized variational graph autoencoder (ARVGA)~\cite{Pan2018}, the encoder is forced to generate embedding vectors resembling the data from a prior distribution through adversarial training. Incorporating an attention mechanism into convolutional neural networks (CNNs) \cite{Zheng2022a}  or GNNs~\cite{Xia2022} is another technique that has been exploited. These techniques have also been jointly applied for specific applications such as graph link prediction~\cite{Weng2020}, graph clustering visualization~\cite{Weng2020}, and node classification in partially labeled graphs~\cite{Xiao2023}.

Generally in  recent studies based on GNNs for image retrieval, input images are represented as the graph nodes whose features are usually extracted by pre-trained CNNs 
to construct the graph. However, the exploited pre-trained CNN models are typically trained on natural images based on the ImageNet dataset~\cite{Deng2010}, and their extracted features may not be optimally suited for medical data (e.g., histological images)~\cite{Kipf2016, denner2024leveraging}. On the other hand, recently, several pre-trained foundation models, trained on millions of data samples in an unsupervised manner, have been developed and applied to a wide range of computer vision tasks.  
	
In this study, we adapted and exploited attention-based adversarially regularized variational graph autoencoders (A-ARVGAE) for the histological breast image retrieval task. Unlike most prior studies that employ conventional pre-trained CNNs for feature extraction to construct the graph, we used medical foundation models, including contrastive language-image pre-training with medical data (BioMedCLIP)~\cite{zhang2023large}, general-purpose self-supervised model for computational pathology (UNI)~\cite{Chen2024}, and clustering-guided contrastive learned model (CCL)~\cite{Wang2023}, as the graph node feature extractors. Evaluated on two binary and multi-class histopathological breast image datasets, namely the breast cancer histopathological database (BreakHis)~\cite{Spanhol2016} and the breast cancer histology image dataset (BACH)~\cite{Aresta2019}, our results demonstrate the superior performance of models trained with foundation model features over those with CNN features, with UNI delivering the best overall retrieval performance. The main contribution of this research can be summarized as follows:



\begin{itemize}
    \item We adapted and exploited one of the state-of-the-art GNN models, which incorporated both attention mechanism and adversarial training for the histology image retrieval task of breast cancer.
    \item Instead of using conventional pre-trained CNN features as the node feature extractor, we made use of medical foundation model features. 
    \item Through extensive experiments, we show the excellent performance of our proposed approach for both utilized GNN and feature extraction on two histopathological breast image datasets.
\end{itemize}

The rest of this paper is organized as follows: In Section~\ref{sec:Related Works}, we present a brief background on methods for medical image retrieval and graph representation learning in histopathology. Section~\ref{sec:Methodology} describes our methodology, which contains all steps for the proposed framework. Section~\ref{sec:Results} consists of the experiments and evaluations with an analysis of the results. Finally, Section~\ref{sec:Conclusion} concludes the paper with suggestions for future works.

\section{Related Works}
\label{sec:Related Works}
This section contains a brief review of medical image retrieval and graph representation learning in histopathology, considering their relevance to our work.
		
\subsection{Medical Image Retrieval}	
Various methods have been proposed in the literature for CBMIR. 
CBMIR models rely on the content (features) of the images and can be used unsupervised without the need for manual labeling. 
Conventional CBMIR models consist of image preprocessing, hand-crafted or automatic feature extraction, and feature similarity measurements using different distance-based methods. 

Commonly used preprocessing methods include noise reduction, contrast enhancement, patch generation from whole slide images (WSIs), and normalization (e.g., stain normalization in histological images)~\cite{10250462, app122211375}. However, many novel machine learning (ML)- and deep learning (DL)-based feature extractors do not require any preprocessing steps. 

For feature extraction, both manually designed and more recently DL-based automatic feature extractors have been utilized. 
Many pre-trained DL-based models, such as EfficientNet~\cite{Tan2021}, and DenseNet121~\cite{8099726}, have been employed as feature extractors for various computer vision tasks including medical image retrieval~\cite{Ukwuoma2022, mahbod2021pollen}. Hand-crafted or automatically extracted features of a query image are then compared to the image features in a database to find similar ones. For this comparison, various similarity measurements such as distance-based retrieval (SDR), customized query approach (CQA), or nearest-neighbor matching have been used~\cite{Suju2017}. In the context of histopathology, Yottixel~\cite{Kalra2020} and FISH~\cite{Chen2022} methods select the most distinctive patches to represent WSIs through $K$-means clustering based on RGB histogram and spatial coordinate features. Yottixel employs a pre-trained DenseNet121, trained on the ImageNet dataset, to extract patch embeddings, which are then transformed into binary barcodes. FISH trains a model based on an autoencoder on the cancer genome atlas (TCGA) dataset to encode an integer index for each patch during retrieval. However, it is worth mentioning that using a feature extractor trained only on natural images based on the ImageNet dataset might be sub-optimal due to the fundamental differences between natural and medical/histopathological images. 

With the advent of transformer-based models, the availability of large-scale unlabeled data and advanced vision-based or vision-language-based training methods such as contrastive language-image pre-training (CLIP)~\cite{pmlr-v139-radford21a} or knowledge distillation with no Labels (DINO)~\cite{9709990}, several foundation models have been developed and pre-trained on large scale data in an unsupervised or self-supervised manner. These foundation models have been used for various applications such as off-the-shelf feature extractors for computer vision tasks and include general-purpose foundation models such as OpenCLIP or medical foundation models such as UNI (for histopathological images)~\cite{Chen2024} or PanDerm (for dermatoscopic images)~\cite{yan2024general}. In this work, in addition to using well-known CNN model features for benchmarking, we employed three medical foundation models as graph node feature extractors and compared their results. The foundation models utilized include BioMedCLIP (a medical foundation model trained on diverse medical data types) and UNI and CCL (where both are medical foundation models specifically trained on histological images).


	
\subsection{Graph Representation Learning in Histopathology}
While CNNs have demonstrated impressive performance in medical image analysis, they often have trouble at capturing contextual information from neighboring areas, especially with a limited field of view. In contrast to CNNs, GNNs are better at preserving complex correlations among neighboring elements during the learning process. 

In the field of computational pathology, several GNN models, such as cell graphs, tissue graphs, cell-tissue graphs, and patch graphs, have been proposed for tasks like node classification (e.g., nuclei classification) or graph classification (e.g., histologic image patch categorization)~\cite{Anand2020, Ahmedt-Aristizabal2021a}. However, these applications have primarily been applied in a supervised manner and often need advanced preprocessing. Using other models like HoVerNet~\cite{graham2019hover} or DDU-Net~\cite{10.3389/fmed.2022.978146} to generate nuclei segmentation masks as input for the graph-based model is an example of such preprocessing. 
As highlighted in a recent review~\cite{Ahmedt-Aristizabal2021}, most GNN-based works in computational pathology are related to supervised classification tasks. Only a few studies have investigated histological image retrieval~\cite{Wang2023, Ahmedt-Aristizabal2021, Zheng2018, 10.1007/978-3-030-32239-7_61}. However, even in these studies, the potential of attention mechanisms and adversarial training have not been explored. In our research, for the first time, we adapted and used both attention mechanism and adversarial training in variational graph autoencoders for the histological image retrieval task.



\section{Methodology}
\label{sec:Methodology}
\begin{figure}
    \centering
    \includegraphics[width=1.0\linewidth]{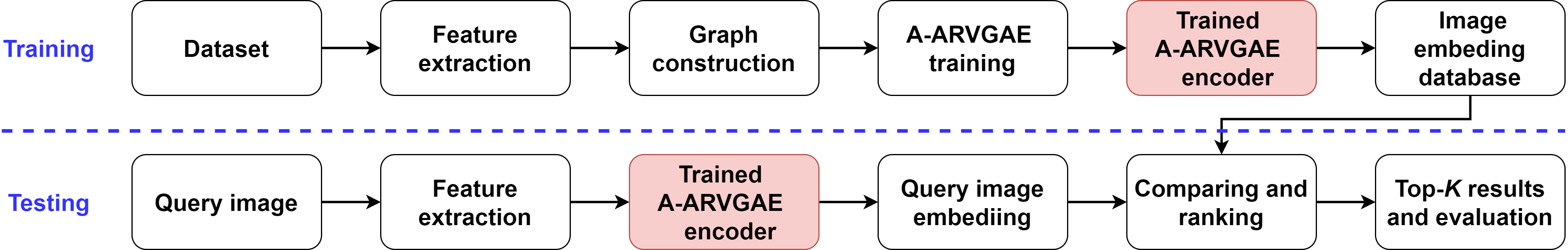}
    \caption{The proposed general workflow of the breast histological image retrieval model involves distinct processes for training and testing. A-ARVGAE: Attention-based adversarially regularized variational graph autoencoders. }
    \label{fig:workflow}
\end{figure}

The general workflow of our proposed approach for histological image retrieval is depicted in Fig.~\ref{fig:workflow}. In the following, we discuss the different parts of the workflow in detail.

\subsection{Dataset}
\begin{figure}[t!]
    \centering
    \begin{tabular}{ccccc}
    \footnotesize		 BreakHis(40$\times$) & \footnotesize	 BreakHis(100$\times$)  &  \footnotesize		 BreakHis(200$\times$) & \footnotesize	 BreakHis(400$\times$) & \footnotesize	 BACH(200$\times$)\\
    \includegraphics[width=2.2cm]{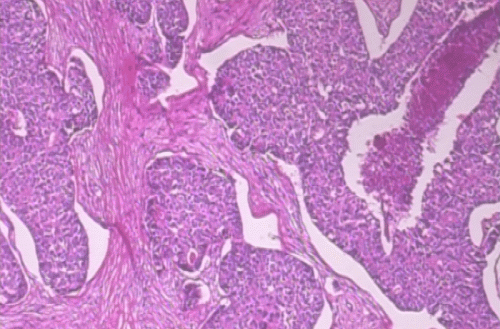} &
    \includegraphics[width=2.2cm]{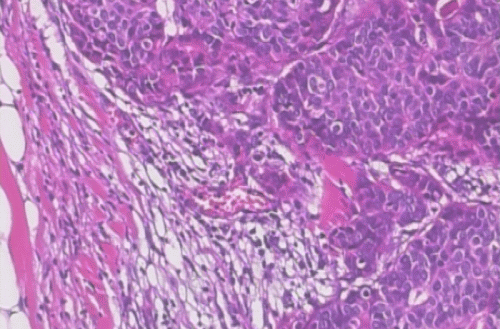} &
    \includegraphics[width=2.2cm]{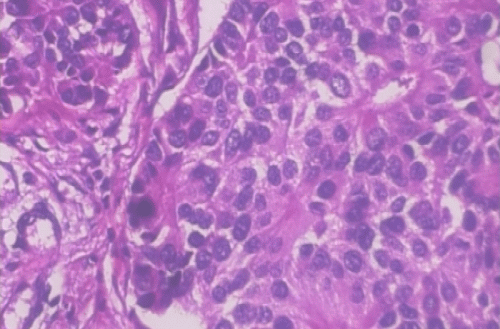} &
    \includegraphics[width=2.2cm]{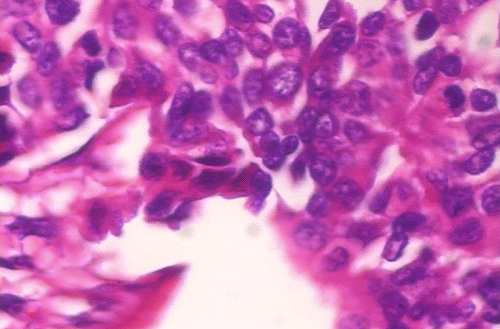} &
    \includegraphics[width=2.2cm]{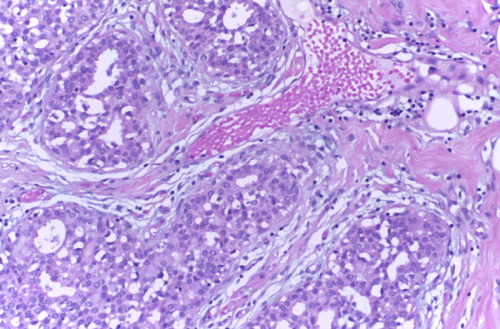} 
    \end{tabular}
    \caption{Example images from the BreakHis (first four images) and the BACH (last image) datasets.}
    \label{fig:examples}
\end{figure}

We use two breast histopathology image datasets in this study, namely the BreakHis and the BACH dataset, which are subsequently detailed. Example images from the datasets are shown in Fig.~\ref{fig:examples}.

\begin{itemize}
    \item \textbf{BreakHis:} This dataset contains hematoxylin and eosin (H\&E)-stained breast histology microscopy images captured at various magnifications ($40\times$, $100\times$, $200\times$, $400\times$) and is categorized into two primary classes, including benign and malignant classes. The dataset contains 7,909 images of which are 2,480 benign and 5,429 malignant samples with a size of $700\times460$ pixels in Portable Network Graphics (PNG) format.
	
    \item \textbf{BACH:} This dataset contains H\&E-stained breast histology microscopy images captured at $200\times$ magnification. It includes 400 images distributed across four distinct classes, namely benign (100 images), in situ carcinoma (100 images), invasive carcinoma (100 images), and normal (100 images), each with a resolution of $2048\times1536$ pixels in Tagged Image File (TIF) format.	
\end{itemize}

\subsection{Feature Extraction}
The first step in our proposed retrieval model is the feature extraction of all images in the dataset. Instead of using hand-crafted features or conventional pre-trained CNNs as feature extractors, we utilize pre-trained medical foundation models as the main feature extractors. 

Unlike conventional pre-trained CNNs, the foundation models in this study are specifically trained on medical/histological images. We leverage three medical foundation models in this study, namely BioMedCLIP, UNI and CCL. BioMedCLIP is a vision-language model based on the CLIP architecture and is initially trained on the PMC15M dataset, which contains approximately 15 million image-text pairs extracted from PubMed medical articles. It utilizes a vision transformer (ViT) as its image encoder and a bidirectional encoder representation from transformers (BERT) model as its text encoder. In this study, we use only the image encoder for feature extraction. UNI is a vision model trained with a self-supervised DINO V2~\cite{oquab2023dinov2} approach on over 100 million histological image patches sourced from 100,000 WSIs. UNI uses ViT-based models in student and teacher branches and has shown superb performance across various computer vision tasks. CCL is also a vision model, trained using a novel self-supervised, clustering-guided contrastive learning technique on over 15 million histological image patches. The CCL approach integrates two contrastive losses (a weighted InfoNCE and a group-level InfoNCE). It utilizes a clustering-guided memory bank to optimize positive and negative sample identification during contrastive learning. Further details on these models can be found in their respective studies~\cite{zhang2023large, Chen2024, Wang2023}. We hypothesize that these models, trained on vast datasets of medical images, will yield superior features for histological image retrieval.

We also use standard baseline models as feature extractors and compare their retrieval performance with the foundation model feature extractors. For this purpose, we employ well-known pre-trained CNN models, namely VGG19~\cite{Simonyan2015}, MobileNetV2~\cite{Ekoputris2018}, DenseNet121~\cite{8099726}, 
NASNetLarge~\cite{Zoph2018}, and EfficientNetV2M~\cite{Tan2021}. 



\subsection{Graph Construction}
GNNs are mainly applied to graph-based datasets. These datasets typically include nodes, node features, edges, and adjacency matrices. In contrast, non-structural datasets, such as image-based data, lack inherent graph structures. This makes graph construction an essential step for applying GNNs to such data. For the retrieval task, graph construction methods often treat subjects (e.g., images) as graph nodes, with the features of these subjects acting as node features. Edges between nodes can be defined using distance metrics like cosine and Euclidean distances, indicating the likelihood of interaction between entities~\cite{Lin2023, Gao2022}. By utilizing the edges and nodes, the adjacency matrix can be constructed where the values of the matrix act as indicators, quantifying the connections between nodes.

The topology of the graphs is usually determined experimentally, employing strategies such as a pre-defined proximity threshold, a nearest neighbor rule, or a probabilistic model. Various algorithms, such as $k$ approximate nearest neighbor ($k$-ANN), can create edges between nodes. This approach is more time-efficient than the $k$-NN algorithm and beneficial, especially in high-dimensional feature spaces, where the complexity of high dimensions slows down exact nearest neighbor searches. While there are various implementations of ANN algorithms, the ANN benchmark~\cite{Aumuller2017} is a widely used library encompassing most ANN-based methods. In this study, we utilize the fast library for approximate nearest neighbors (FLANN) to calculate distances between node embeddings, selecting the $k$-nearest nodes to each entity as adjacent nodes. 
	
\begin{figure}
    \centering
    \includegraphics[width=0.7\linewidth]{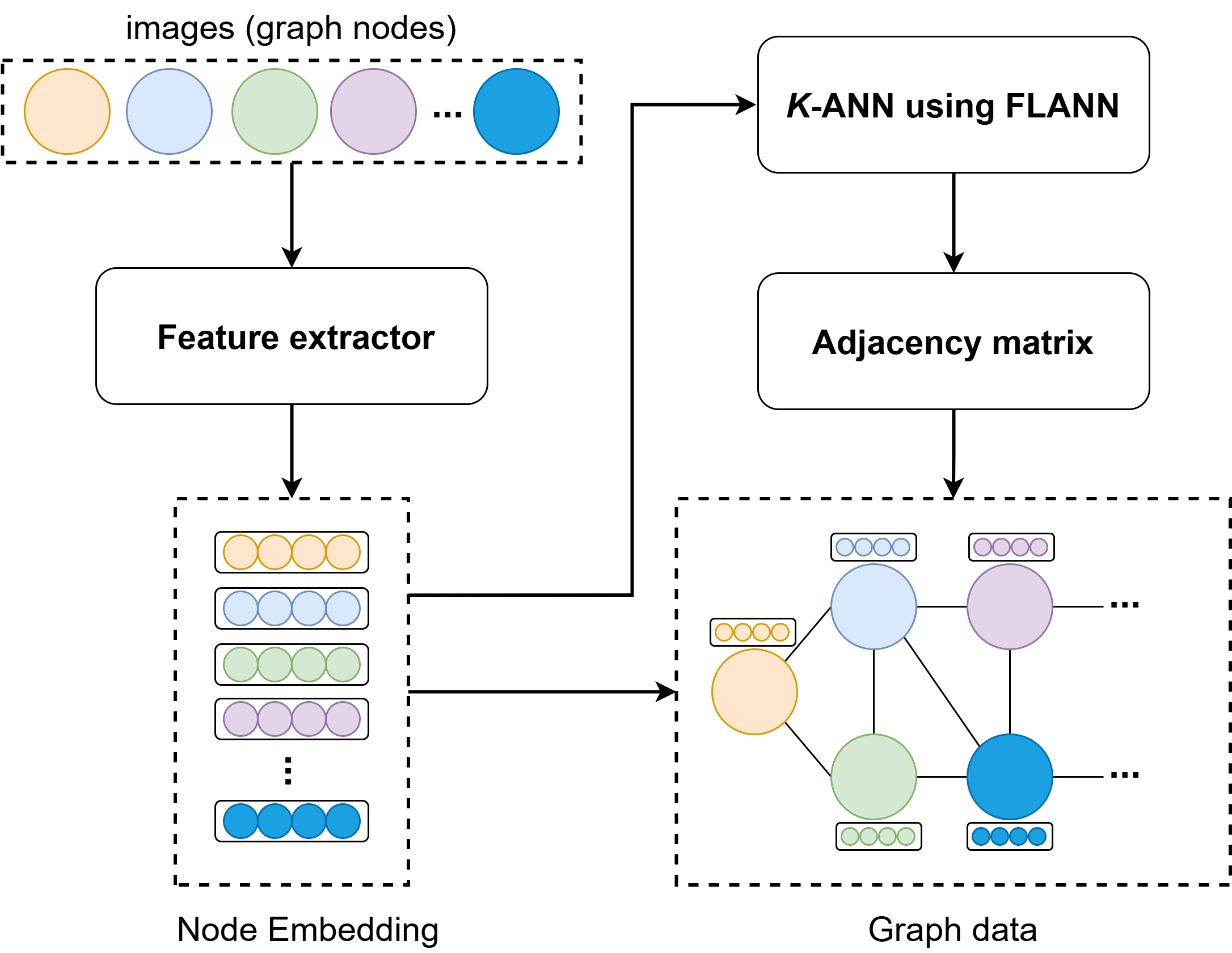}
    \caption{The framework of graph construction. $K$-ANN: $K$ approximate nearest neighbors, FLANN: fast library for approximate nearest neighbors  }
    \label{fig:GNN_construction}
\end{figure}

We use transductive learning, where the entire dataset (training, validation, and test data) is observed to build the graph. The summarized workflow of building the graph is shown in Fig.~\ref{fig:GNN_construction}. 
	
\subsection{A-ARVGAE}
\begin{figure}
    \centering
    \includegraphics[width=1\linewidth]{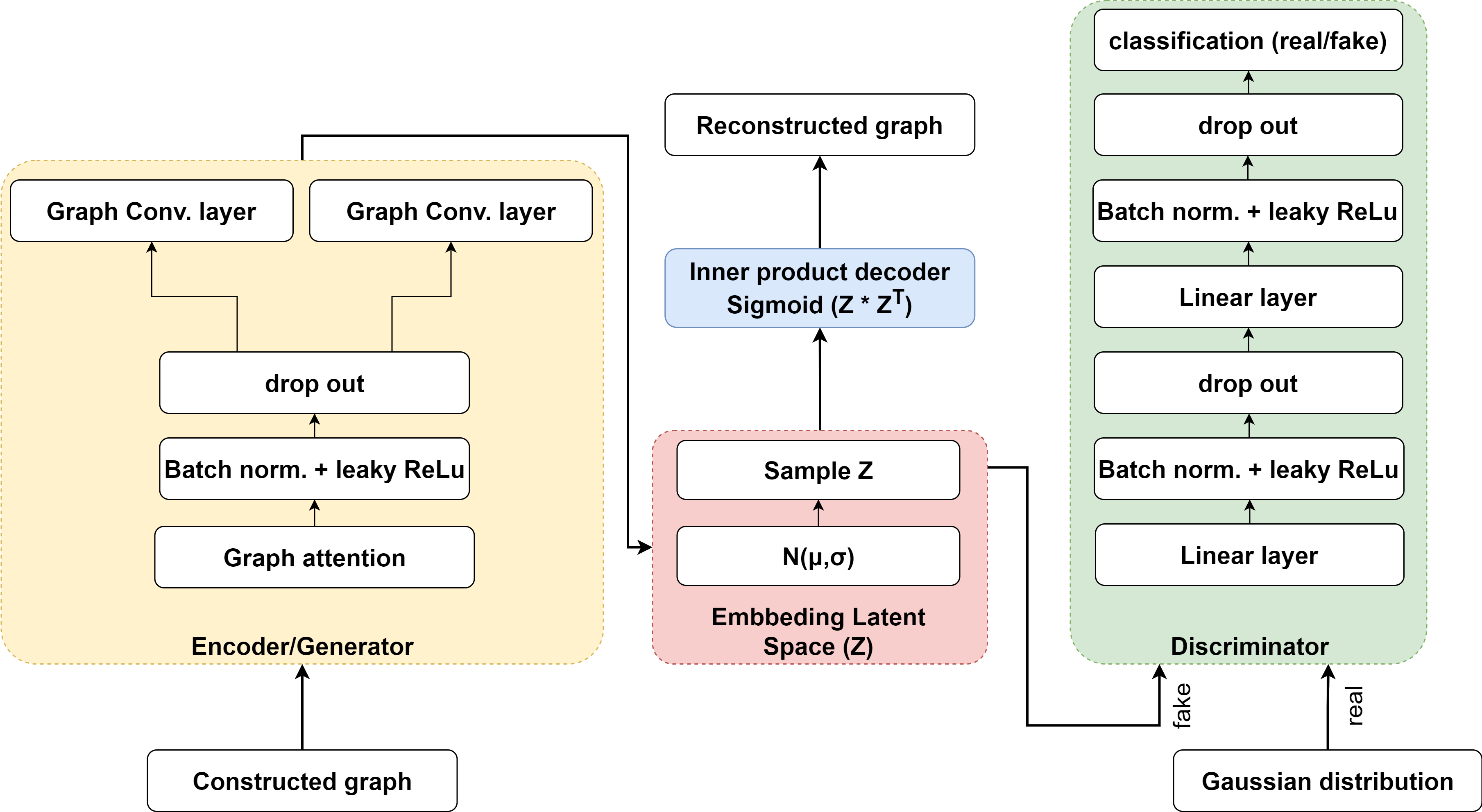}
    \caption{The Architecture of attention-based adversarially regularized variational graph autoencoder (A-ARVGAE). For interpretation of the references to color in this figure legend, the reader is referred to the web version of this article.}
    \label{fig:GNN_model}
\end{figure}

The overall architecture of the A-ARVGAE model is depicted in Fig.~\ref{fig:GNN_model}. The proposed architecture comprises four main components: the encoder (yellow block), which also serves as the generator for adversarial training, the embedding latent space (pink block), the decoder (blue block), and the discriminator (green block), which is another essential component for adversarial training. 
 	
The constructed graph, as described in the previous section, is fed into the encoder of the A-ARVGAE model. The encoder includes a graph attention layer that performs attention operations, batch normalization, leaky ReLU, and dropout, followed by two graph convolutional layers. The attention-based convolutional operation integrates attention mechanisms into convolutional layers by assigning varying weights to the neighbors and enabling the model to concentrate on the most relevant parts of the graph when learning node representations~\cite{Velickovic2017}. 
As indicated by prior studies, a multi-head attention layer has yielded superior performance~\cite{Weng2020}; hence, we also employ the two-head attention layer in our work.
	
Inspired by the VGAE~\cite{Kipf2016}, the encoder in our architecture does not encode an input into a fixed latent representation. Instead, the outputs of the graph convolutional layers are parameters (mean ($\mu$) and variance ($\sigma$)) that define a probability distribution over the possible latent representations, denoted as $Z$, characterized by the Normal distribution $N(\mu, \sigma)$. Sampling from the latent space involves drawing samples from the distribution $N(\mu, \sigma)$ to produce latent node embeddings. This process employs the reparameterization trick, as introduced in~\cite{Kipf2016}, which enables the backpropagation of gradients through the stochastic sampling process for gradient descent optimization.
	 
The decoder block employs a linear projection through a straightforward inner product ($Z * Z^{T}$) followed by a sigmoid function, which results in the reconstructed graph.
	 
As the A-ARVGAE model incorporates adversarial training, it includes a generator and a discriminator. Within this framework, the encoder functions analogously to a generator in an adversarial network~\cite{Pan2018}. The generator tries to deceive the discriminator by producing synthetic data, which, in this context, refers to latent variables generated from the encoding of graph data. On the other hand, the discriminator (green block) attempts to differentiate whether samples are derived from real data or the generator. The discriminator evaluates embeddings from the prior distribution (a Gaussian distribution) as real and those from the latent variable $Z$ as synthetic or fake. The discriminator block comprises a straightforward MLP with two linear layers along with batch normalization, leaky reLU, and dropout to distinguish between real (a prior Gaussian distribution) and fake (latent space $Z$) embeddings.

The complete model integrates three loss components: a regression-based reconstruction loss (associated with VGAE), a Kullback–Leibler divergence loss (associated with VGAE), and a binary cross-entropy loss (related to discriminator in adversarial training). Further details regarding the model architecture and cost functions can be found in the respective studies~\cite{Pan2018, Velickovic2017}.

As an ablation study, we compare the results of our employed GNN model (A-ARVGAE) with other GNN-based models, namely GAE, VGAE, and ARVGA. 
The results of these comparisons are reported in Section~\ref{sec:Results} (Table~\ref{tab:res_gnn}).
	  	
\subsection{Trained Graph Encoder}
After training, the trained A-ARVGAE encoder can transform any input image into a vector in the embedding space.
	
\subsection{Image Embedding Database}
The embeddings generated by the trained A-ARVGAE encoder in the previous step for each training image are stored in a database. This database is then utilized in the testing phase to identify the top-$K$ similar images for a given query image as explained in Section~\ref{sec:Testing phase}.
    
\subsection{Testing phase}
\label{sec:Testing phase}
As depicted in the lower part of Fig.~\ref{fig:workflow}, the model aims to identify similar top-$K$ images in the database for a query image. Similar to the process with training images, features are first extracted from the test image using the pre-trained feature extractors. These features are then fed into the trained A-ARVGAE model to obtain the embedding in a lower-dimensional space, specifically from the encoder part of the model. After acquiring the embedding of the query image, it is compared with the embeddings in the image database to identify similar images. To find the most similar images, the ranking algorithm evaluates and ranks the candidate images based on Euclidean distance. These images are then arranged and ranked in ascending order according to Euclidean distances. Once ranked, similar top-$K$ images are chosen as the final output of the retrieval model.
	
To evaluate the performance of the developed approach, we use two widely-employed evaluation metrics for image retrieval, namely, mean Average Precision (mAP ($k$))~\cite{Zheng2022} and mean Majority Vote (mMV ($k$))~\cite{Chen2022}, where $k$ represents the top k results. 







\section{Results \& Discussion}
\label{sec:Results}
In this section, we start with the implementation details, then delve into the experimental setup, and finally present the obtained results along with a detailed discussion.
			
\subsection{Implementation Details}
We implemented the graph neural network using the PyTorch deep learning framework and the PyTorch Geometric library. We used the Keras framework for feature extraction with pre-trained CNN models, where the feature extraction from BioMedCLIP, UNI, and CCL models was performed using their respective repositories on Hugging Face and GitHub~\footnote{\url{https://huggingface.co/microsoft/BiomedCLIP-PubMedBERT_256-vit_base_patch16_224}\\\url{https://github.com/mahmoodlab/UNI}\\\url{https://github.com/Xiyue-Wang/RetCCL}}

Adam optimizer was used to train the models with an initial learning rate (LR) of 0.0001. For dropout layers, a probability of 0.2 was chosen in the model architecture. All models were trained for 250 epochs, but for the models with adversarial components, the discriminator was trained for five iterations in each epoch. All images were down-sampled to 256$\times$256 pixels to fit into the utilized GPU memory to extract features from the training and test images. For each experiment, an identical training set (70\%), validation set (10\%), and test set (20\%) were used to ensure fair comparisons between different experiments. The data splitting was done randomly by keeping the ratio between different classes in the training, validation, and test sets. In the graph construction process, the parameter $k$ was set to 25 for the BreakHis dataset and 15 for the BACH dataset. For the attention layer in the A-ARVGAE model, a 2-head attention layer was used. All experiments were conducted on a single machine using an NVIDIA GTX 1060 GPU and an Intel Core™ i7-9750H CPU.
	
\subsection{Experiments}
\noindent We performed two sets of experiments for each dataset in this study. 
In the first set, we investigated the effect of using the medical foundation model feature extractors against other pre-trained CNN models for the histology image retrieval task for breast cancer. The results of this experiment for the BreakHis and BACH datasets are reported in Table~\ref{tab:res_breakhis}, Table~\ref{tab:res_breakhis_avg}, and Table~\ref{tab:res_bach}. For the BreakHis dataset, Table~\ref{tab:res_breakhis} shows the results per image magnification, while Table~\ref{tab:res_breakhis_avg} provides the average results across magnifications. For a fair comparison, we used the A-ARVGAE model in all experiments.

\begin{table}[!hp]
		\caption{Evaluation of the retrieval performance for the BreakHis dataset with different feature extractors across different magnifications. The best results for convolutional neural networks and foundation models are highlighted by underlining and bold font, respectively. The attention-based adversarially regularized variational graph autoencoder is used in all experiments as the retrieval model.
		}
		\label{tab:res_breakhis}
		\centering
		\begin{tabular}{|l|l|c|c|}
			\hline
			\textbf{Magnification} & \textbf{Feature Extraction} & \textbf{mAP(5)} & \textbf{mMV(5)} \\
			\hline
			$40\times$ & VGG19    & 0.945             & 0.859 \\
			& MobileNetV2         & \underline{0.957} & \underline{0.894} \\
			& DenseNet121         & 0.941             & 0.864 \\
			& NASNetLarge         & 0.925             & 0.809 \\
			& EfficientNetV2M     & 0.919             & 0.768 \\ \cline{2-4}
			& BioMedCLIP          & 0.983             & 0.934 \\
			& UNI                 & \textbf{0.985}    & \textbf{0.954} \\
			& CCL                 & 0.978             & 0.949 \\
			\hline
			$100\times$ & VGG19   & 0.948             & 0.850 \\
			& MobileNetV2         & 0.947             & \underline{0.884} \\
			& DenseNet121         & \underline{0.953} & 0.879 \\
			& NASNetLarge         & 0.919             & 0.777 \\
			& EfficientNetV2M     & 0.915             & 0.826 \\ \cline{2-4}
			& BioMedCLIP          & 0.979             & 0.951 \\
			& UNI                 & 0.973             & 0.946 \\
			& CCL                 & \textbf{0.980}    & \textbf{0.961} \\
			\hline
			$200\times$ & VGG19   & 0.941             & 0.840 \\
			& MobileNetV2         & \underline{0.955} & \underline{0.870} \\
			& DenseNet121         & 0.946             & 0.860 \\
			& NASNetLarge         & 0.921             & 0.776 \\
			& EfficientNetV2M     & 0.905             & 0.791 \\ \cline{2-4}
			& BioMedCLIP          & 0.968             & 0.925  \\
			& UNI                 & \textbf{0.975}    & 0.910  \\
			& CCL                 & 0.964             & \textbf{0.930} \\
			\hline
			$400\times$ & VGG19   & 0.945             & 0.861 \\
			& MobileNetV2         & 0.941             & 0.850 \\
			& DenseNet121         & \underline{0.955} & \underline{0.895} \\
			& NASNetLarge         & 0.906             & 0.751 \\
			& EfficientNetV2M     & 0.905             & 0.790 \\ \cline{2-4}
			& BioMedCLIP          & 0.935             & \textbf{0.861} \\
			& UNI                 & 0.938             & 0.850 \\
			& CCL                 & \textbf{0.940}    & 0.839 \\
			\hline

		\end{tabular}
\end{table}

\begin{table}[!hp]
	\caption{Evaluation of the average retrieval performance for the BreakHis dataset with different feature extractors.  The best results for convolutional neural networks and foundation models are highlighted by underlining and bold font, respectively. The attention-based adversarially regularized variational graph autoencoder is used in all experiments as the retrieval model.  
	}
	\label{tab:res_breakhis_avg}
	\centering
	\begin{tabular}{|l|l|c|c|}
		\hline
		\textbf{Magnification} & \textbf{Feature Extraction} & \textbf{mAP(5)} & \textbf{mMV(5)} \\
		\hline

		BreakHis Avg.& VGG19 & 0.944             & 0.852 \\
		& MobileNetV2        & \underline{0.950} & \underline{0.874} \\
		& DenseNet121        & 0.948             & \underline{0.874} \\
		& NASNetLarge        & 0.917             & 0.778 \\
		& EfficientNetV2M    & 0.911             & 0.791 \\ \cline{2-4}
		& BioMedCLIP         & 0.966             & 0.917\\
		& UNI                & \textbf{0.967}    & 0.915 \\
		& CCL                & 0.965             & \textbf{0.919} \\
		\hline
	\end{tabular}
\end{table}
	
\begin{table}[!htp]
	\caption{Evaluation of the retrieval performance for the BACH dataset with different feature extractors. The best results for convolutional neural networks and foundation models are highlighted by underlining and bold font, respectively. The attention-based adversarially regularized variational graph autoencoder is used in all experiments as the retrieval model.}
	\label{tab:res_bach}
	\centering
	\begin{tabular}{|l|l|c|c|}
		\hline
		\textbf{Magnification} & \textbf{Feature Extraction} & \textbf{mAP(5)} & \textbf{mMV(5)} \\
		\hline
		$200\times$ & VGG19   & \underline{0.899} & \underline{0.787} \\
		& MobileNetV2         & 0.893             & 0.737 \\
		& DenseNet121         & 0.897             & 0.700 \\
		& NASNetLarge         & 0.892             & 0.800 \\
		& EfficientNetV2M     & 0.882             & 0.762 \\ \cline{2-4}
		& BioMedCLIP          & 0.942             & 0.846  \\
		& UNI                 & \textbf{0.976}    & \textbf{0.942} \\
		& CCL                 & 0.947             & 0.913 \\
		\hline
	\end{tabular}
\end{table}
	
As observed from Table~\ref{tab:res_breakhis}, for the BreakHis dataset, models trained with foundation model features outperform all models trained with CNN features across both metrics, except for 400$\times$ magnification, where DenseNet121 delivered better retrieval performance.  As shown in Table~\ref{tab:res_breakhis_avg}, for the average performance across all magnifications, the employed medical foundation models deliver superior performances compared with all other CNN models, including DenseNet121. Based on the average results for the BreakHis dataset, the top-performing foundation model achieves a 1.7\% and 4.5\% improvement over the best CNN model in terms of mAP(5) and mMV(5), respectively. 

The results in Table~\ref{tab:res_bach} also indicate the superior performance of the medical foundation model feature extractors compared with other CNN models for the BACH dataset. Based on the results for the BACH dataset, the top-performing foundation model achieves a 7.7\% and 15.5\% improvement over the best CNN model in terms of mAP(5) and mMV(5), respectively.
	
Overall, the results presented in Table~\ref{tab:res_breakhis_avg} and Table~\ref{tab:res_bach} for the BreakHis and BACH datasets indicate that feature extractors trained on medical/histological images outperform those trained on non-medical images. Additionally,
comparing foundation models shows highly competitive performance across all foundation models on the BreakHis dataset, while UNI achieves the best performance on the BACH dataset.

In the second experiment, we compare the performance of the employed A-ARVGAE model with other GNN-based models for both BreakHis and BACH datasets. The results of these experiments are presented in Table~\ref{tab:res_gnn}. The UNI model was used as the graph feature extractor to derive the results in this table. 

\begin{table}[!htp]
	\caption{Comparison of different GNN-based models for the BreakHis and BACH datasets for the breast cancer histological image retrieval task. The UNI model was used as the graph feature extractor in all experiments. GAE: graph autoencoder, VGAE: variational graph autoencoder, ARVGA: adversarial regularized variational graph autoencoder, A-ARVGAE: attention-based adversarially regularized variational graph autoencoder. }
	\label{tab:res_gnn}
	\centering
	\begin{tabular}{|l|l|c|c|}
		\hline
		\textbf{Dataset} &   \textbf{Method} & \textbf{mAP(5)} & \textbf{mMV(5)} \\
		\hline
		BreakHis (40$\times$)  &GAE      & 0.897           & 0.783\\
		 & VGAE                          & 0.960           & 0.894\\
		 & ARVGA                         & 0.930           & 0.817\\
		 & A-ARVGAE                      & \textbf{0.985}  & \textbf{0.954} \\
		\hline
		BreakHis (100$\times$) &GAE      & 0.896           & 0.768\\
		 & VGAE                          & 0.960           & 0.937\\
		 & ARVGA                         & 0.949           & 0.922\\
		 & A-ARVGAE                      & \textbf{0.973}  & \textbf{0.946}\\
		\hline
		BreakHis (200$\times$) &GAE      & 0.907           & 0.771\\
		 & VGAE                          & 0.930           & 0.855\\
		 & ARVGA                         & 0.938           & 0.865 \\
		 & A-ARVGAE                      & \textbf{0.975}  & \textbf{0.910}\\
		\hline
		BreakHis (400$\times$) &GAE      & 0.925           & 0.784 \\
		 & VGAE                          & 0.934           & 0.790 \\
		 & ARVGA                         & 0.930           & 0.817\\
		 & A-ARVGAE                      & \textbf{0.938}  & \textbf{0.850} \\
		\hline
		BreakHis Avg.          &GAE      & 0.906           & 0.776 \\
		 & VGAE                          & 0.946           & 0.869 \\
		 & ARVGA                         & 0.936           & 0.855\\
		 & A-ARVGAE                      & \textbf{0.967}  & \textbf{0.915} \\
		\hline
		BACH (200$\times$)     &GAE      & 0.941           & 0.884 \\
		 & VGAE                          & 0.962           & 0.865 \\
		 & ARVGA                         & 0.962           & 0.855 \\
		 & A-ARVGAE                      & \textbf{0.976}  & \textbf{0.942}\\
		\hline
	\end{tabular}
\end{table}

As shown in Table~\ref{tab:res_gnn}, the A-ARVGAE model surpasses all GNN-based models, i.e., GAE, VGAE, and ARVGA of both datasets. The superior performance of the proposed model over ARVGA demonstrates the significance of the attention component in the employed architecture. An attention component allows the model to focus on specific parts of input data to calculate the model weights. The average results indicate that the A-ARVGAE model achieves 2.1\%/4.6\% and 1.4\%/5.8\% higher performance in terms of mAP and mMV compared to the second-best GNN model for the BreakHis and BACH datasets.
	
	

\begin{figure}[t!]
    \centering
    \begin{tabular}{cc|ccccc}
    &	 Query & 	Rank 1   &  	Rank 2		 &  	Rank 3	  & 	Rank 4	&  	Rank 5 \\ \hline

		 \begin{turn}{90}\footnotesize BH (40$\times$)\end{turn} &\includegraphics[width=1.7cm]{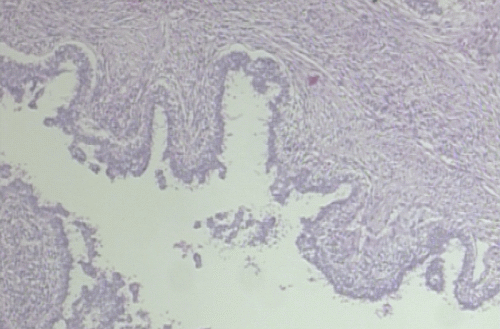} &
		\includegraphics[width=1.7cm]{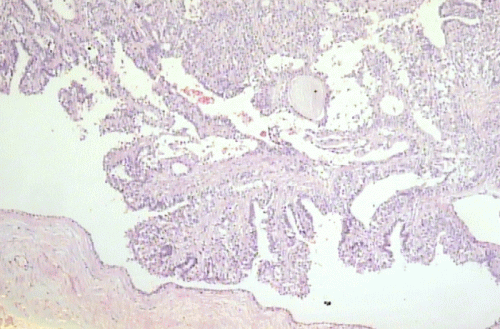} &
		\includegraphics[width=1.7cm]{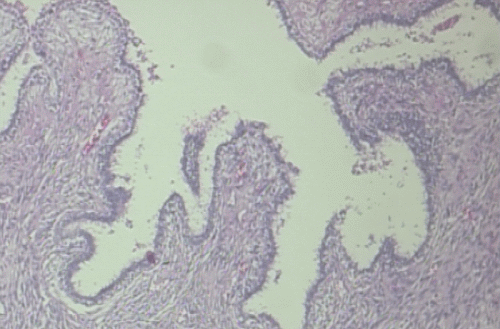} &
		\includegraphics[width=1.7cm]{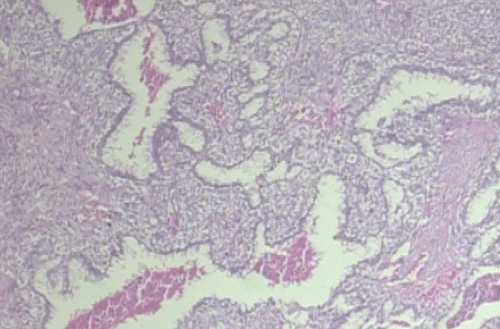} &
		\includegraphics[width=1.7cm]{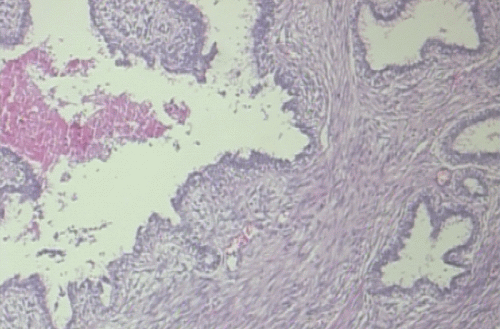} &
		\includegraphics[width=1.7cm]{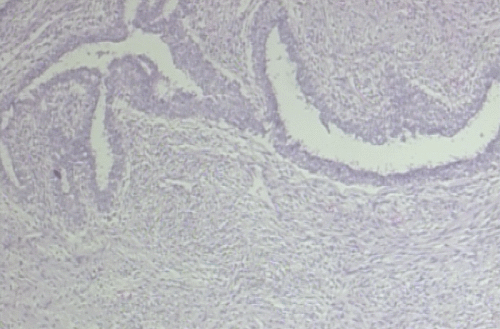} \\
		  & BN & M \color{red}\xmark& BN \color{green}\cmark & BN \color{green}\cmark & BN \color{green}\cmark & BN \color{green}\cmark \\
		  \hline
		  
		  		 \begin{turn}{90} \footnotesize BH (100$\times$)\end{turn} &\includegraphics[width=1.7cm]{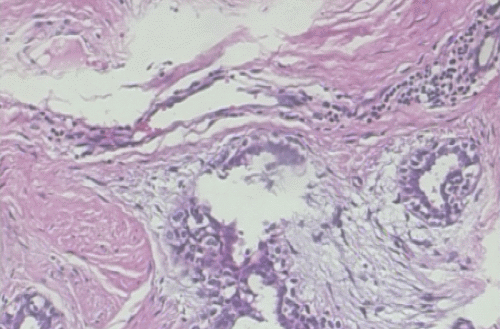} &
		  \includegraphics[width=1.7cm]{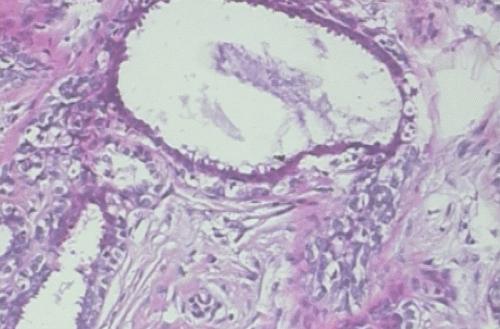} &
		  \includegraphics[width=1.7cm]{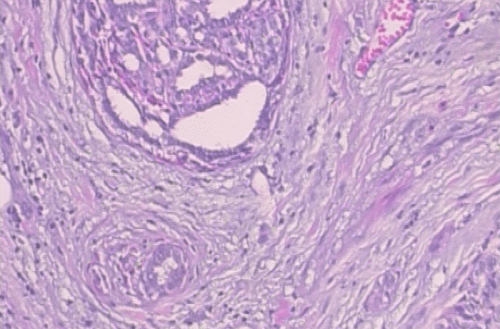} &
		  \includegraphics[width=1.7cm]{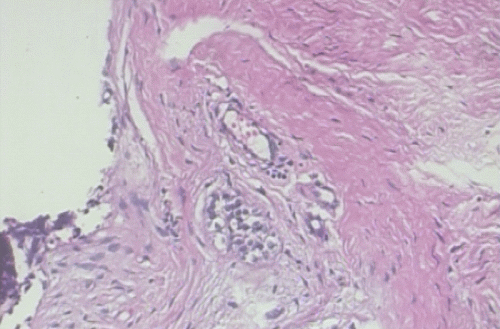} &
		  \includegraphics[width=1.7cm]{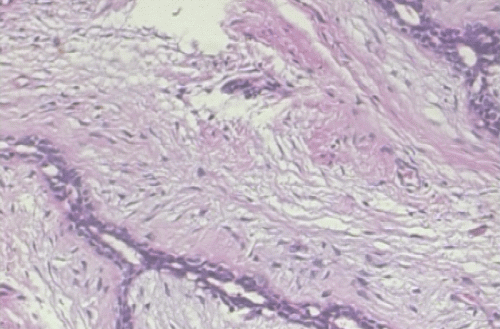} &
		  \includegraphics[width=1.7cm]{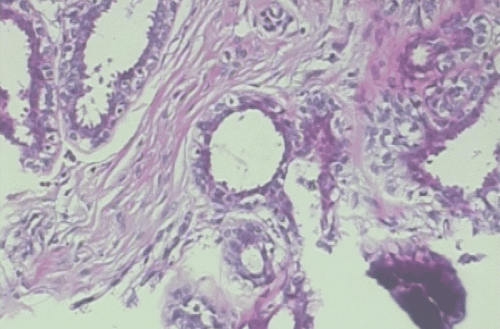} \\
		  & BN & BN \color{green}\cmark& M \color{red}\xmark & BN \color{green}\cmark & BN \color{green}\cmark & BN \color{green}\cmark \\
		  \hline
		  
		  		 \begin{turn}{90} \footnotesize BH (200$\times$)\end{turn} &\includegraphics[width=1.7cm]{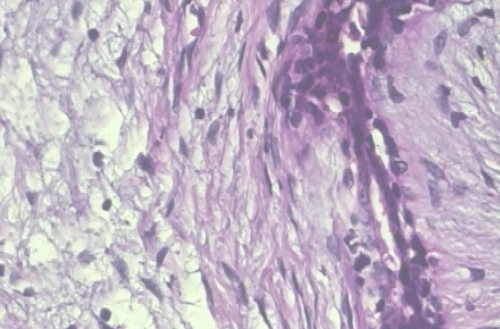} &
		  \includegraphics[width=1.7cm]{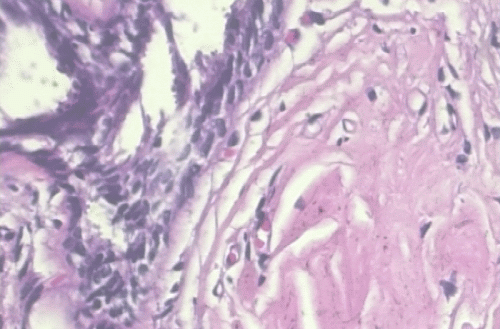} &
		  \includegraphics[width=1.7cm]{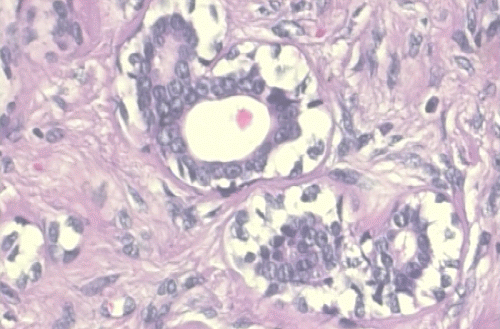} &
		  \includegraphics[width=1.7cm]{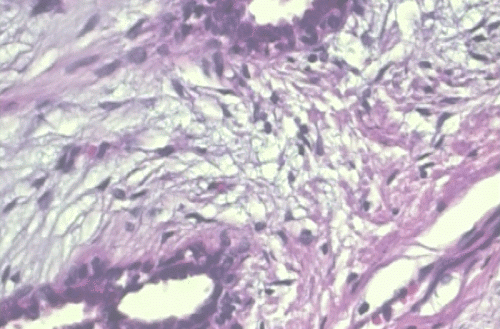} &
		  \includegraphics[width=1.7cm]{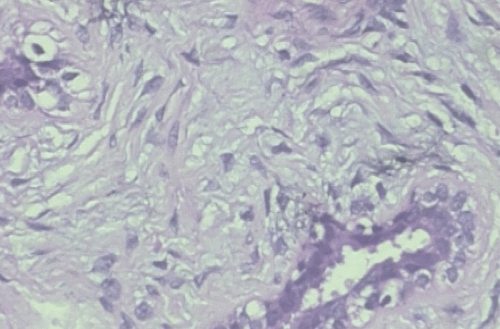} &
		  \includegraphics[width=1.7cm]{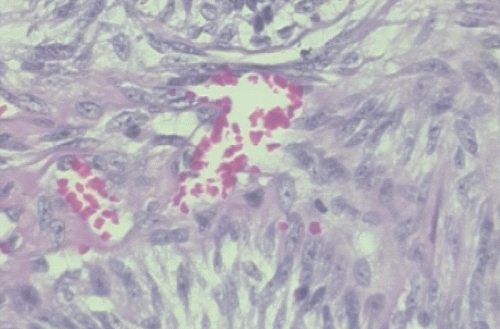} \\
		  & BN & BN \color{green}\cmark & BN \color{green}\cmark & BN \color{green}\cmark & BN \color{green}\cmark & BN \color{green}\cmark \\
		  \hline
		  
		  		 \begin{turn}{90} \footnotesize BH (400$\times$)\end{turn} &\includegraphics[width=1.7cm]{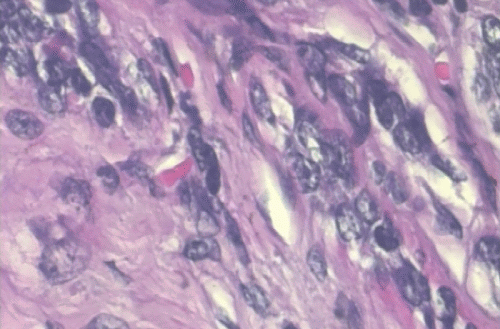} &
		  \includegraphics[width=1.7cm]{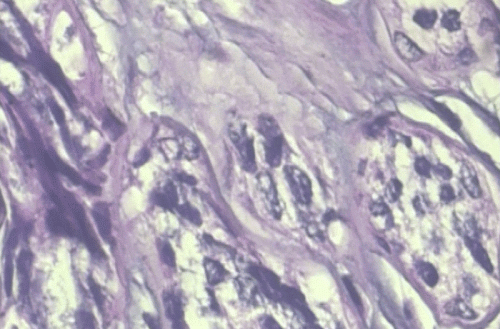} &
		  \includegraphics[width=1.7cm]{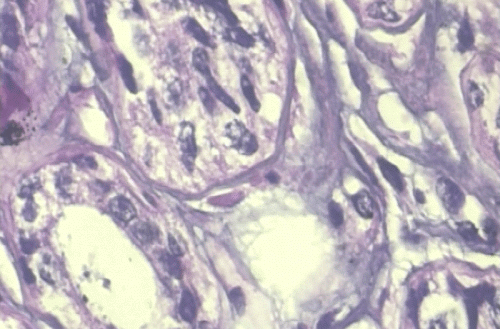} &
		  \includegraphics[width=1.7cm]{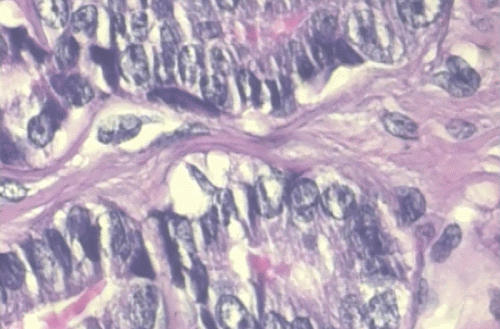} &
		  \includegraphics[width=1.7cm]{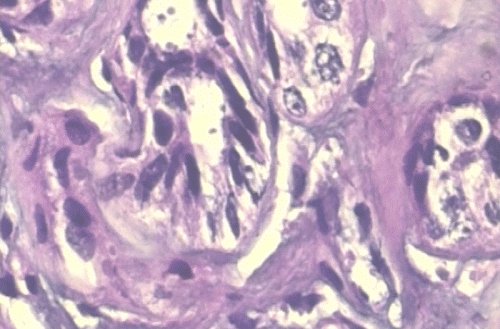} &
		  \includegraphics[width=1.7cm]{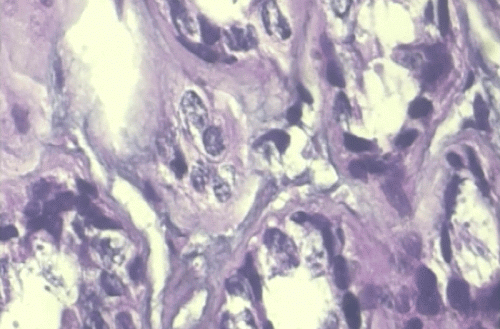} \\
		  & BN & BN \color{green}\cmark & BN \color{green}\cmark & BN \color{green}\cmark & BN \color{green}\cmark & BN \color{green}\cmark \\
		  \hline
		  
		  		 \begin{turn}{90} \footnotesize BACH (200$\times$)\end{turn} &\includegraphics[width=1.7cm]{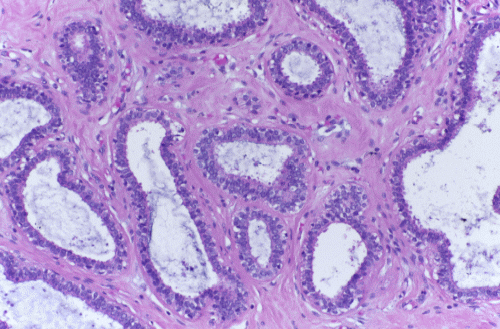} &
		  \includegraphics[width=1.7cm]{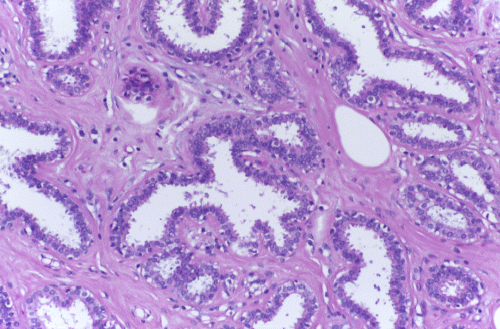} &
		  \includegraphics[width=1.7cm]{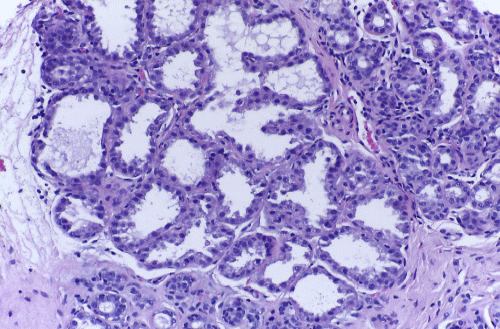} &
		  \includegraphics[width=1.7cm]{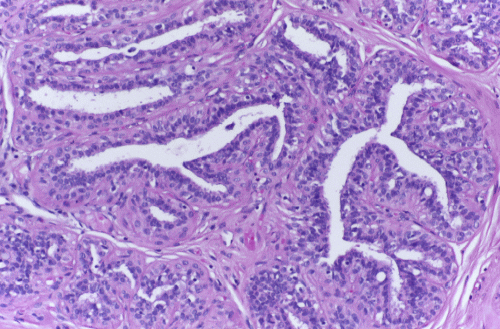} &
		  \includegraphics[width=1.7cm]{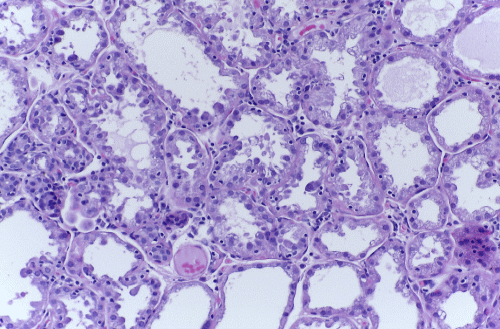} &
		  \includegraphics[width=1.7cm]{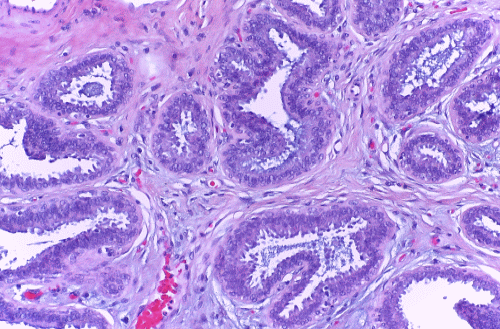} \\
		  & BN & BN \color{green}\cmark& BN \color{green}\cmark & BN \color{green}\cmark& BN \color{green}\cmark & BN \color{green}\cmark \\
		  \hline

	\end{tabular}
	\caption{Five sample queries from the BreakHis and BACH datasets and their top similar retrieved images. Successful and unsuccessful retrieval are denoted by the symbols ({\color{green}\cmark}) and ({\color{red}\xmark}), respectively. M: Malignant class, BN: Benign class, BH: BreakHis}
	\label{fig:ranking}
\end{figure}

For visual analysis, we randomly select five query images from the BreakHis and BACH test sets (four from the BreakHis with different magnifications and one from the BACH dataset) to visualize the model output (top five ranks). Fig.~\ref{fig:ranking} depicts the query images and model output, where the model (A-ARVGAE with UNI feature extractor) provides relevant images (with identical classes acting as the query images) for most samples. 

\begin{figure}[t!]
	\centering
	\begin{tabular}{lll}
		\multicolumn{2}{c}{Similar Pairs}    & \multirow{2}{*}{\includegraphics[width=5cm]{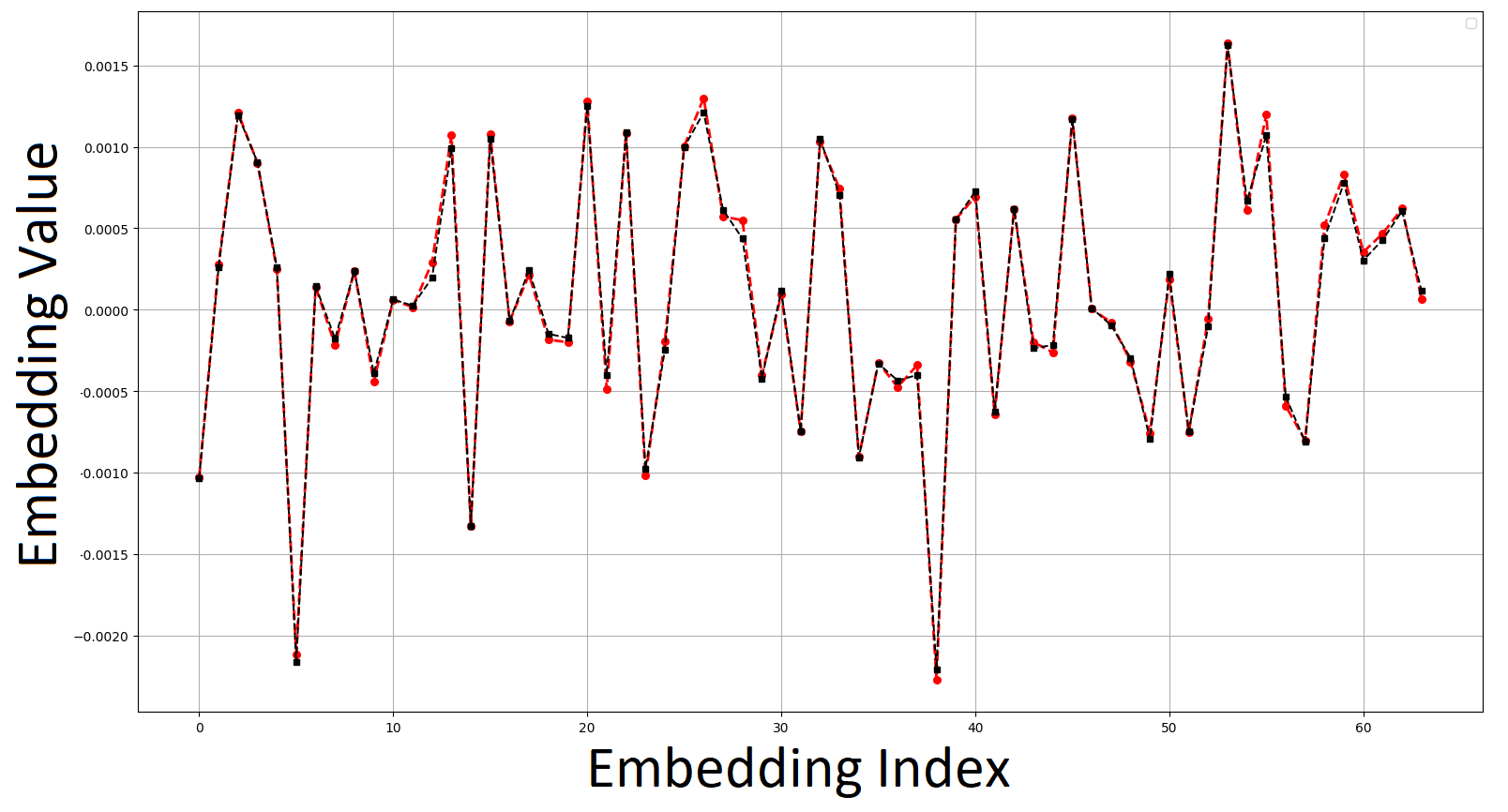}}  \\
		\includegraphics[width=3cm]{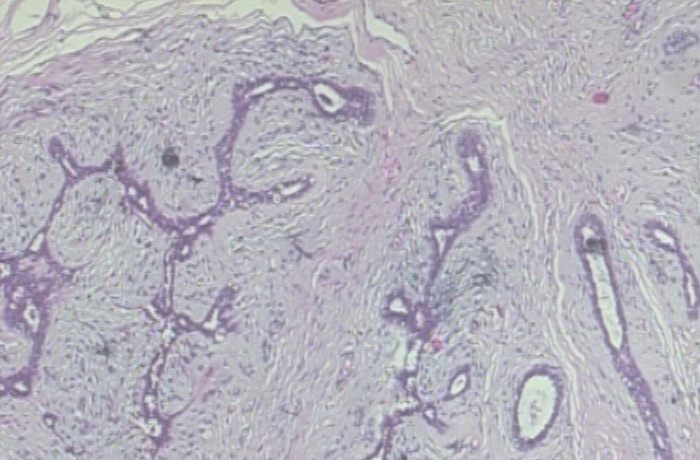}           &                \includegraphics[width=3cm]{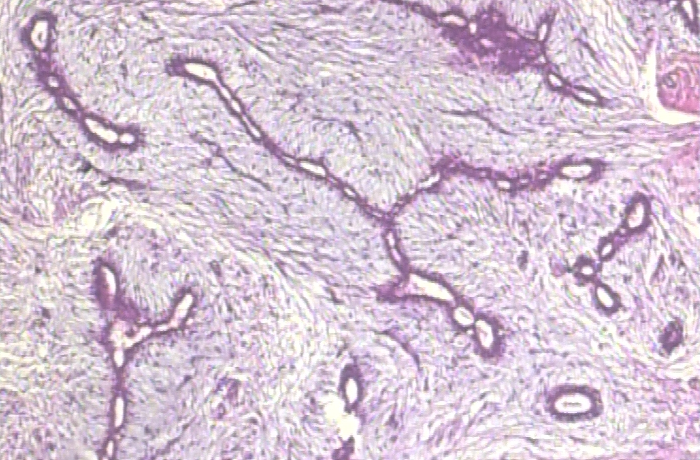}   \\
		\\

			\multicolumn{2}{c}{Dissimilar Pairs}    & \multirow{2}{*}{\includegraphics[width=5cm]{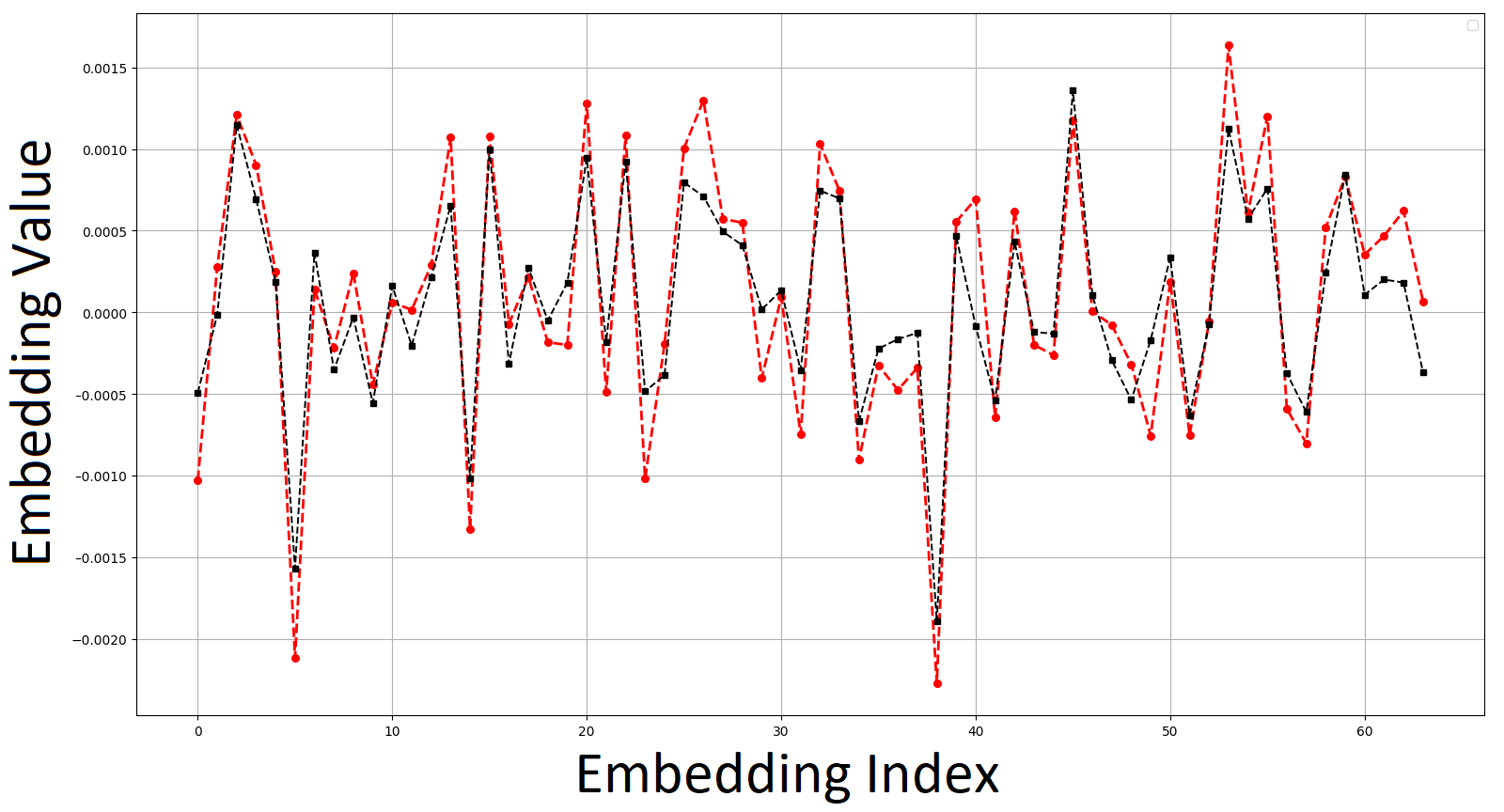}}  \\
		\includegraphics[width=3cm]{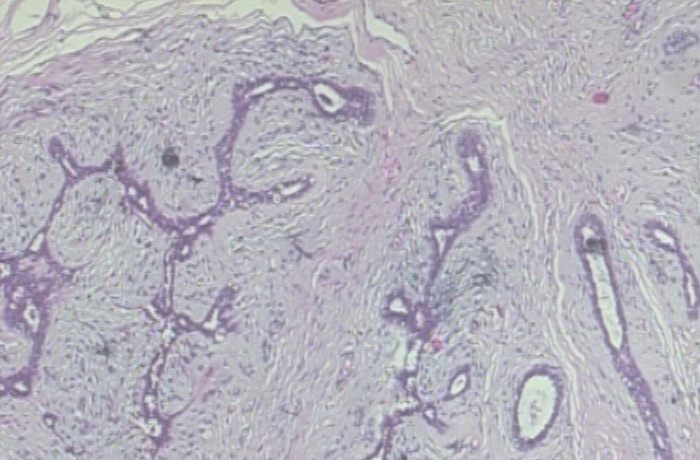}           &                \includegraphics[width=3cm]{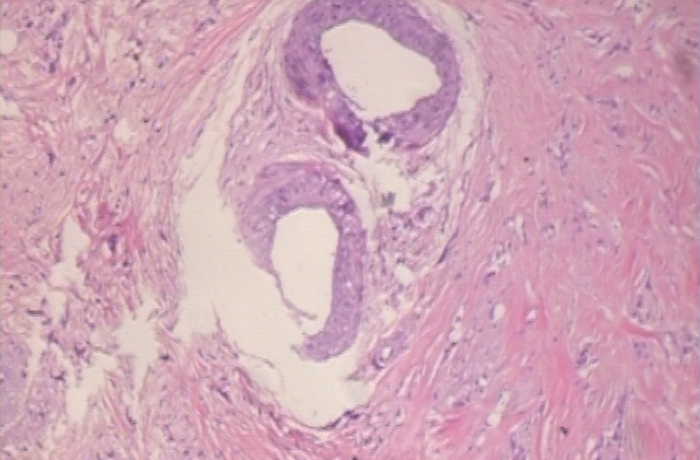}   \\ \\
		
			\multicolumn{2}{c}{Similar Pairs}    & \multirow{2}{*}{\includegraphics[width=5cm]{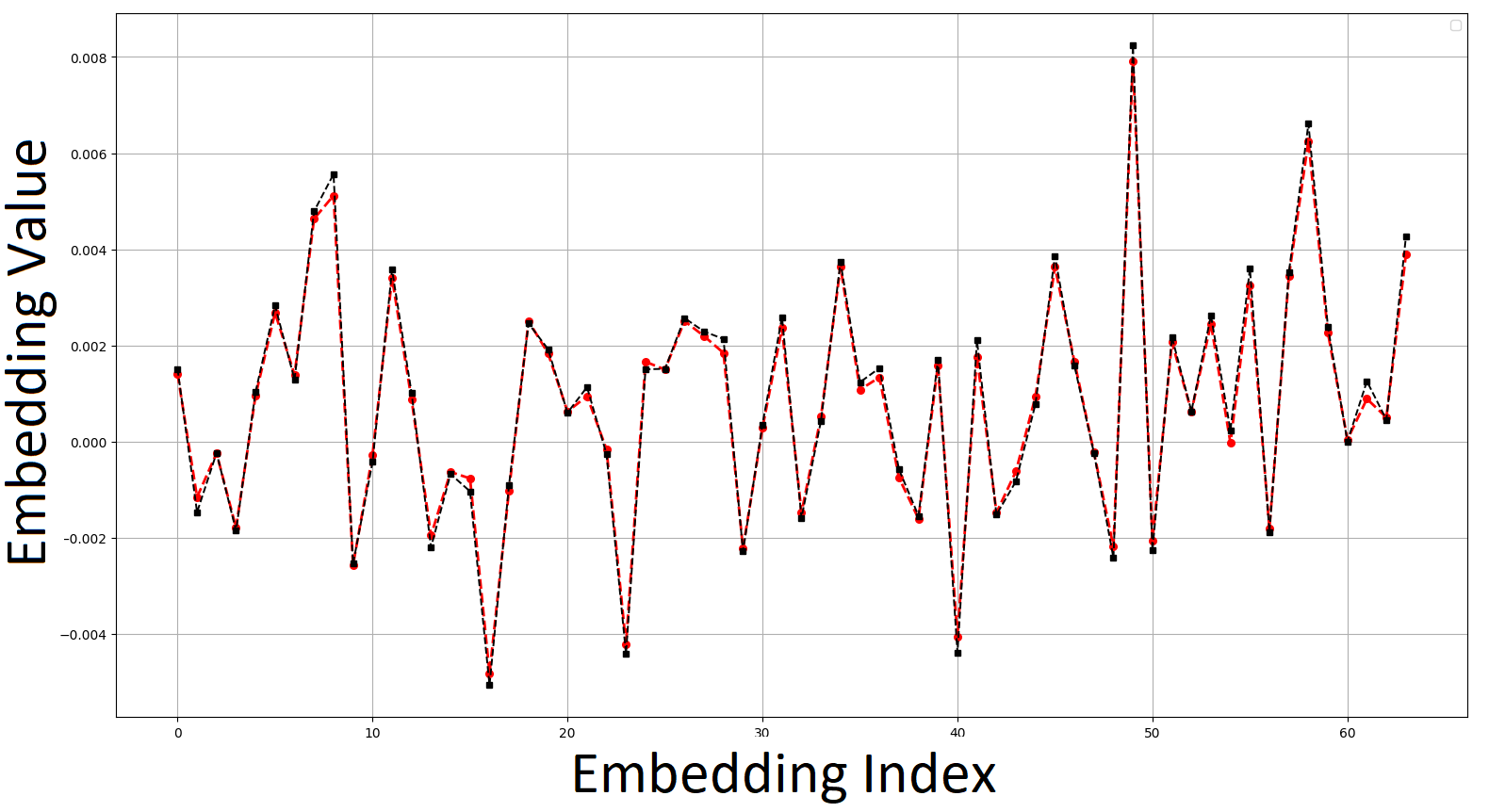}}  \\
		\includegraphics[width=3cm]{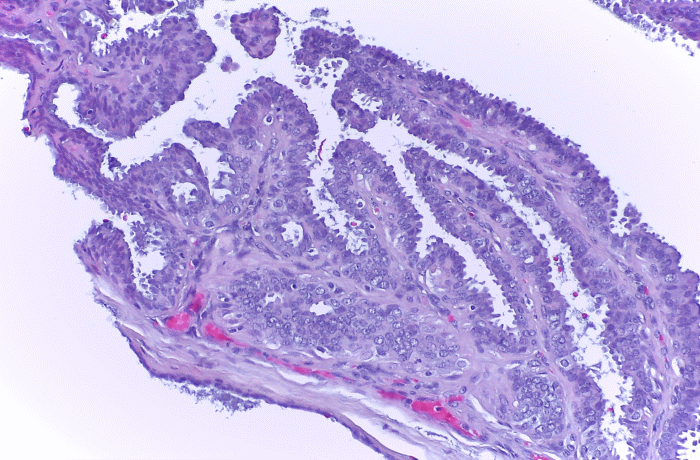}           &                \includegraphics[width=3cm]{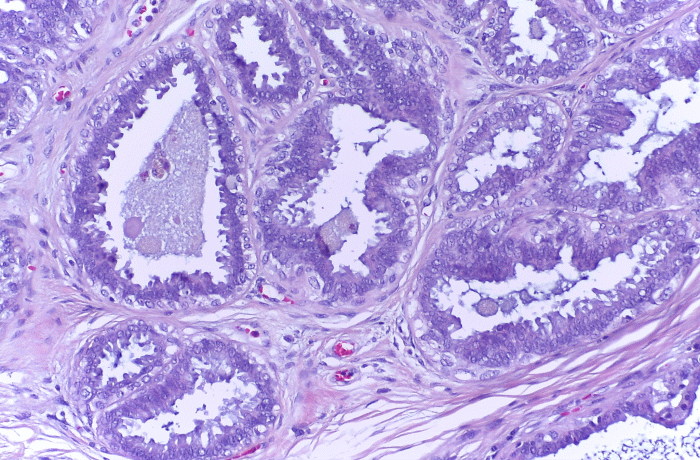}   \\ \\
		
			\multicolumn{2}{c}{Dissimilar Pairs}    & \multirow{2}{*}{\includegraphics[width=5cm]{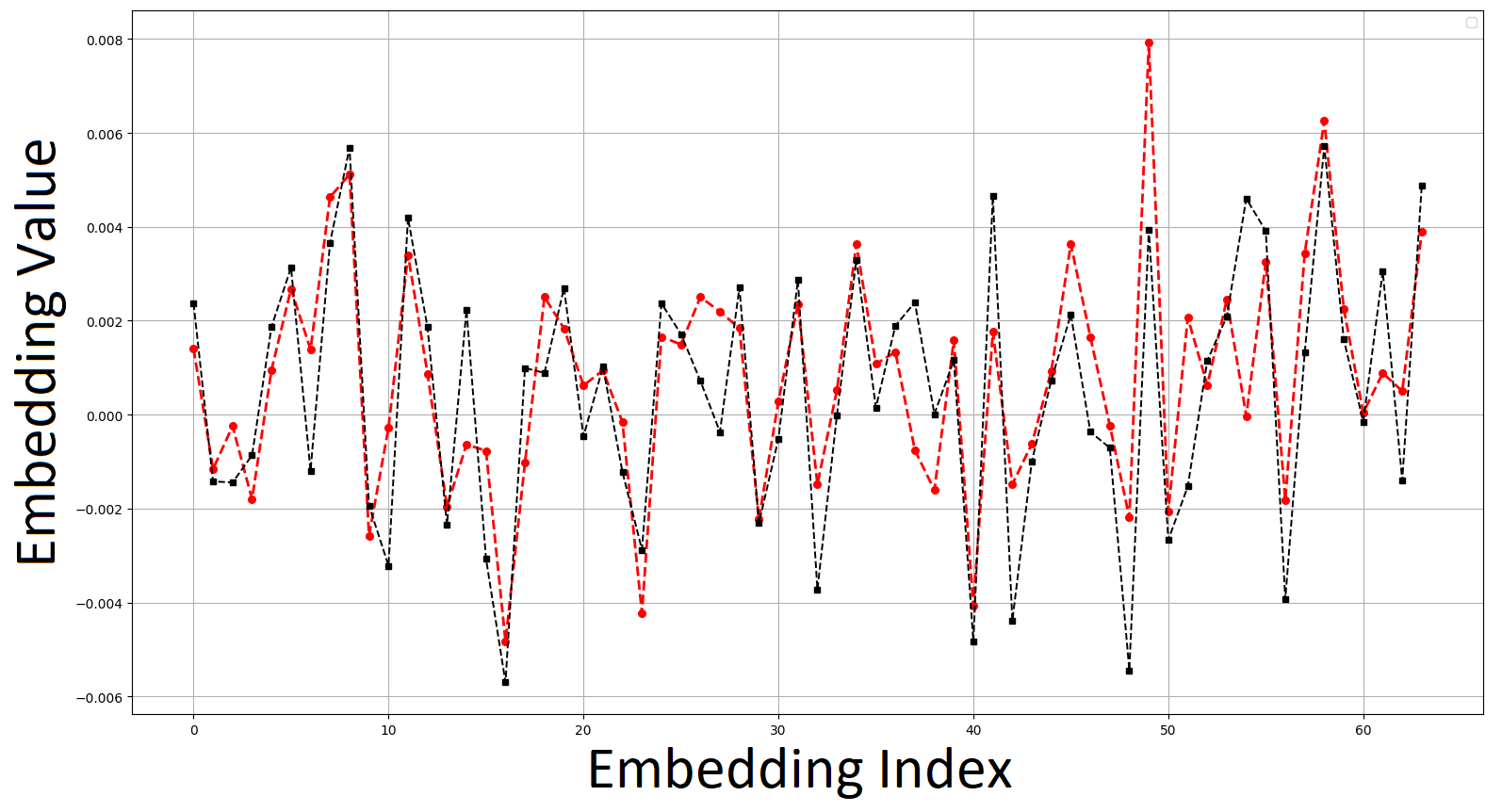}}  \\
		\includegraphics[width=3cm]{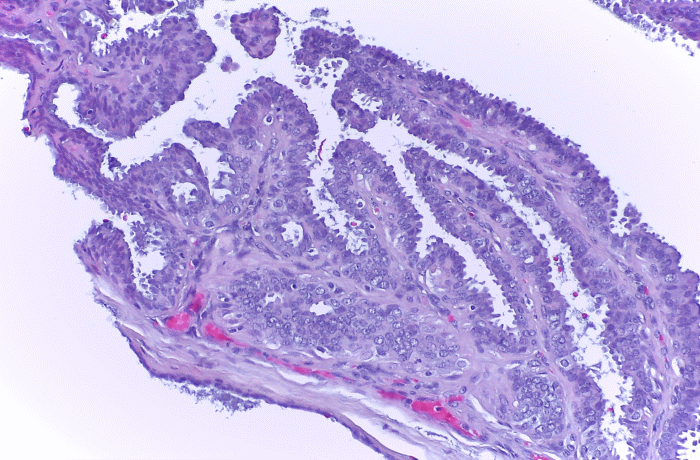}           &                \includegraphics[width=3cm]{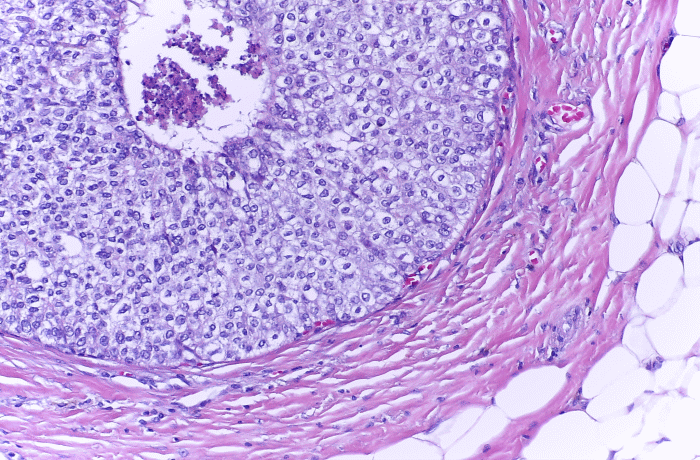}   \\ 
            
	\end{tabular}
		\caption{Comparison of embedding values for four image pairs. Similar pairs refer to images within the same class, and dissimilar pairs refer to images in different classes. The distances between the embedding of the two images in each image pair are shown on the right side of the figure. For each image pair, the left and right image embeddings are depicted in black and red, respectively. }
	\label{fig:embedding}
\end{figure}

As suggested in~\cite{PARCHAM2021115649}, we also compare the embedding vectors of some example image pairs with different degrees of similarity in Fig.~\ref{fig:embedding}, in which the embedding vectors for similar image pairs are very close. In contrast, the embedding vectors for dissimilar pairs are separable.

\section{Conclusion \& Future Works}
\label{sec:Conclusion}
Recently, there has been a growing trend in utilizing GNNs to address challenges in analyzing histopathology images. This research introduces a new GNN-based approach specifically designed for breast histopathology image retrieval. We incorporated attention mechanisms and adversarial training in a variational graph auto-encoder model to enhance retrieval performance. Moreover, instead of using conventional pre-trained CNN models as the node feature extractor, we employed the features from medical foundation models, specifically pre-trained on medical/histological images. Our results confirm incorporating attention mechanisms and adversarial training outperforms the other GNN variants. Additionally, using the feature extracted from foundation models further boosts performance in our experiments. While our approach delivers promising results for breast histology image retrieval, certain aspects can be considered for further investigation in future studies.  

Transformers-based models have shown impressive performance in various computer vision tasks. Some Transformer-based GNNs, such as graph transformer networks~\cite{YUN2022104}, have been developed for tasks such as node classification. Such models can be adapted and employed for medical image retrieval.

While we use FLANN to build the graph in our model, other state-of-the-art ANN methods and libraries, such as the Faiss library~\cite{johnson2019billion} or the HNSW method~\cite{8594636}, can be exploited to enhance performance. However, most of these methods are slower than FLANN ~\cite{Suju2017}.

Another potential enhancement for the model is replacing the inner product decoder with a non-linear neural network, a topic that can be addressed in future studies.

We used three medical foundation models in the study as the graph node feature extractors. However, alternative foundation models could also be explored by considering recent advancements in this field and the availability of other medical and non-medical foundation models. We selected these specific medical foundation models because they have demonstrated superior performance than several other foundation models across various computer vision tasks in former studies ~\cite{denner2024leveraging, Chen2024, mahbod2024evaluating}.

Finally, well-known boosting techniques, such as test-time augmentation or ensembling, can enhance performance for medical image classification or segmentation ~\cite{moshkov2020test, MAHBOD2024669, pmlr-v156-bancher21a}. The resulting retrieval performance can be investigated by incorporating such techniques into the workflow.

\section*{Acknowledgments}
This project is partially supported by the Ernst Mach Grant (reference number: MPC-2023-00569).

\section*{Conflict of Interest}
The authors declare that they have no known competing financial interests or personal relationships that could have appeared to influence the work reported in this paper.

\section*{Credit Authorship Contribution Statement}
Nematollah Saeidi: Conceptualization, Methodology, Investigation, Software, Writing – original draft. Hossein Karshenas: Supervision. Bijan Shoushtarian: Supervision. Sepideh Hatamikia: Supervision. Ramona Woitek: Supervision. Amirreza Mahbod: Conceptualization, Validation, Review \& Editing, Supervision.



\section*{Declaration of generative AI and AI-assisted technologies in the writing process}
During the preparation of this work, the authors used Grammarly (version 1.2.75) and ChatGPT (version 4) to check grammar and spelling and, at times, improve the readability of some sentences. After using these tools, the authors reviewed and edited the content as needed and take full responsibility for the content of the publication.

\bibliographystyle{elsarticle-num}
\bibliography{GNN.bib}
	
\end{document}